\documentclass[runningheads]{llncs}

\usepackage{graphicx}
\usepackage{booktabs}
\usepackage{multirow}
\usepackage[accsupp]{axessibility}  

\usepackage{hyperref}

\usepackage{orcidlink}

\begin{document}

\title{rPPG-SysDiaGAN: Systolic-Diastolic Feature Localization in rPPG Using Generative Adversarial Network with Multi-Domain Discriminator
} 
\titlerunning{rPPG-SysDiaGAN}

\author{Banafsheh Adami\inst{1}\orcidlink{0009-0004-0193-577X} \and
Nima Karimian\inst{1}\orcidlink{0000-0002-4590-7170}}

\authorrunning{B. Adami, N. Karimian}

\institute{West Virginia University, Morgantown,WV }

\maketitle

\begin{abstract}
Remote photoplethysmography (rPPG) offers a novel approach to noninvasive monitoring of vital signs, such as respiratory rate, utilizing a camera. Although several supervised and self-supervised methods have been proposed, they often fail to accurately reconstruct the PPG signal, particularly in distinguishing between systolic and diastolic components. Their primary focus tends to be solely on extracting heart rate, which may not accurately represent the complete PPG signal. To address this limitation, this paper proposes a novel deep learning architecture using Generative Adversarial Networks by introducing multi-discriminators to extract rPPG signals from facial videos. These discriminators focus on the time domain, the frequency domain, and the second derivative of the original time domain signal. The discriminator integrates four loss functions: variance loss to mitigate local minima caused by noise; dynamic time warping loss to address local minima induced by alignment and sequences of variable lengths; Sparsity Loss for heart rate adjustment, and Variance Loss to ensure a uniform distribution across the desired frequency domain and time interval between systolic and diastolic phases of the PPG signal.

\end{abstract}

\section{Introduction}
\label{sec:Intoroduction}
Continuously monitoring vital signs without interrupting patients is crucial in healthcare. Traditional methods (e.g., contact PPG, ECG) limit comfort and hygiene due to skin contact~\cite{makowski2021neurokit2}. The combination of computer vision and deep learning has paved the way for remote photoplethysmography (rPPG), a revolutionary method to measure physiological data without physical contact~\cite{verkruysse2008remote}.
Remote photoplethysmography (rPPG) leverages cameras to capture minute changes in facial color~\cite{poh2010advancements}. These variations translate into an rPPG signal, which acts like a contactless version of the traditional PPG signal. This allows for the measurement of vital signs such as heart rate (HR)~\cite{sabour2021ubfc,shi2019atrial}. Early rPPG research relied on manual methods for processing the signal and extracting key features. However, with the rise of deep learning, researchers have developed supervised and unsupervised techniques. These techniques utilize various network architectures and training methods to estimate rPPG signals directly from facial video data\cite{chen2018deepphys,yu2019remote,sun2022contrast,gideon2021way}.

Current deep learning methods for rPPG excel in estimating heart rate and some aspects of heart rate variability, but struggle to capture the full details of the PPG signal waveform. This includes important features such as systolic and diastolic waves. Although these methods are promising, they need improvement to accurately represent the complete picture of physiological information. To address this limitation, we propose a new deep learning architecture called Swin-AUnet. This approach uses a generative adversarial network (GAN) to create rPPG signals that mimic the shape and features of real PPG signals. Our system analyzes the signal in multiple ways, including its timing, frequency, and rate of change. By considering these different aspects and incorporating physiological knowledge, Swin-AUnet can capture the finer details of the PPG waveform, including the peaks and valleys that correspond to systolic and diastolic blood pressure.

\textbf{It is important to note that our method is supervised, as it relies on ground truth PPG signals during training to guide the learning process and ensure accurate rPPG estimation.}

Our main contributions in this paper are as follows: 1) Our system analyzes the rPPG signal in multiple ways, including its timing (original signal and its rate of change) and frequency content. This comprehensive analysis helps the model create a more accurate representation of the PPG signal. Notably, we're the first to incorporate the second derivative of the PPG signal for rPPG estimation from facial videos.
2) We employ a combination of loss functions in both the time and frequency domains. These functions guide the model during training, ensuring that it prioritizes key aspects such as accurate timing, distribution of information between frequencies, and sparsity (appropriate heart rate). Although using multiple functions might seem complex, each one plays a specific role in achieving a faithful reconstruction.
3) We incorporate Dynamic Time Warping (DTW) as a loss function. This helps align the reconstructed rPPG signal with the real PPG signal, even if there are slight shifts over time. This makes the model more robust to variations in the PPG waveform.
4) We use a wavelet transform to analyze the frequency content of the PPG signal. This allows us to examine both high-frequency and low-frequency components simultaneously, providing a richer understanding of the signal's behavior. The specific wavelet function used (DB4 in our case) is chosen because it closely resembles the patterns observed in PPG signals. This allows it to effectively capture the key features that differentiate a real PPG signal from a remote one. 5) 
Previous rPPG research often focused solely on heart rate, which does not provide a complete picture of cardiovascular health. In contrast, our approach uses a second derivative loss function to capture the shape and features of the PPG waveform. This allows for the reconstruction of crucial details like the systolic and diastolic phases, offering valuable insights into the condition of the blood vessel and other hemodynamic parameters.
\begin{figure*}[!htb]
    \centering
    \includegraphics[width=\textwidth]{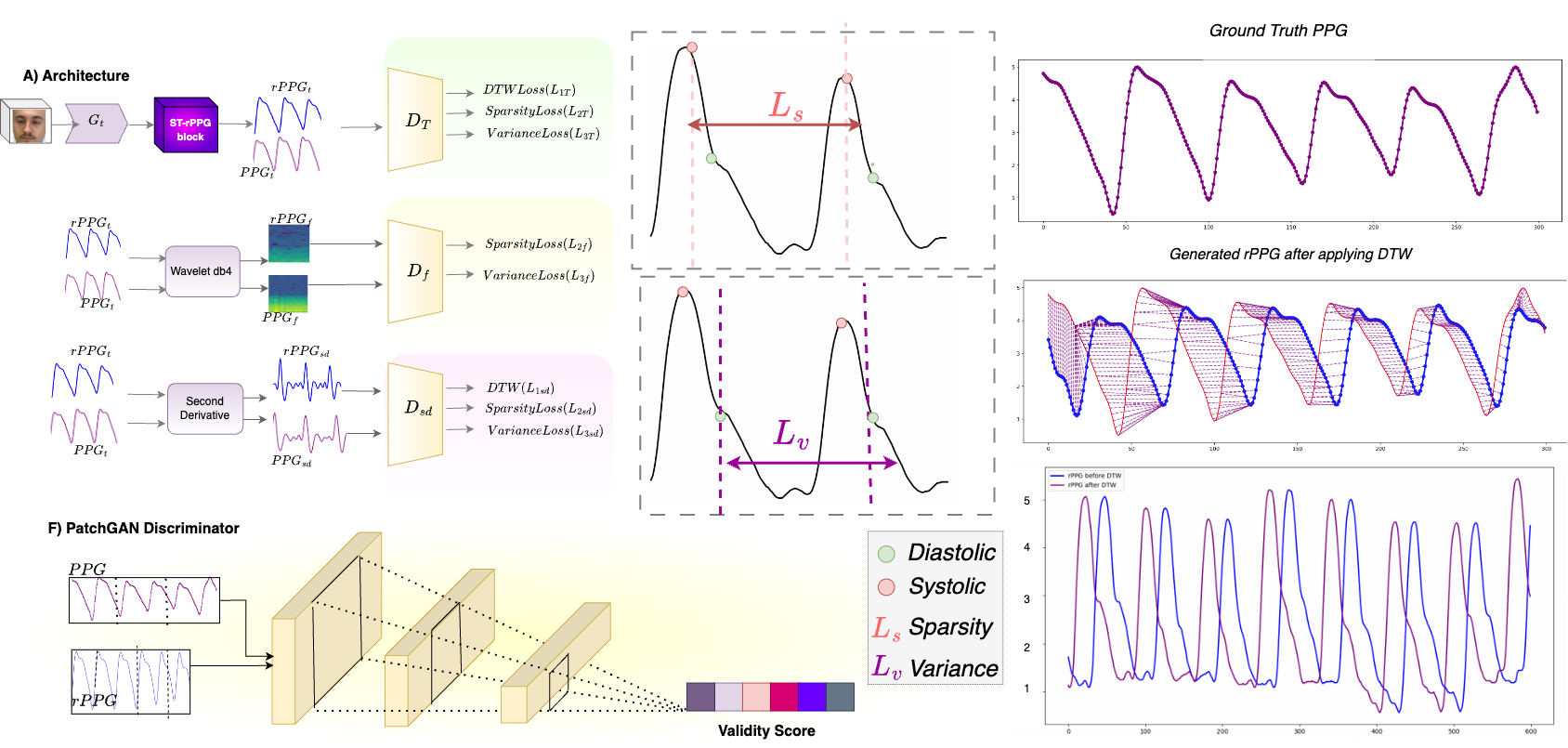}
    \caption{(A) The proposed architecture (GAN) includes one generator and three discriminators: one for the time domain, one for the frequency domain, and one for the second derivative of the time domain signal. (B) Dynamic Time Warping (DTW) illustrates that time series are vertically shifted; however, the ranges of feature values (y-axis values) remain consistent or aligned. (C) Sparsity Loss is used for heart rate adjustment, and Variance Loss ensures a uniform distribution across the desired frequency domain and the time interval between systolic and diastolic phases of the PPG signal.}
    \label{fig:architect}
\end{figure*}

\section{Related Work}
\subsection{Traditional}
Verkruysse et al.~\cite{verkruysse2008remote}  pioneered the use of remote photoplethysmography (rPPG) for physiological measurement in 2008. Since then, various conventional handcrafted methodologies have been developed in the field. Unlike simple averaging of color channels throughout the detected facial area, more effective techniques to extract subtle rPPG signals involve selective integration of information from different color channels~\cite{poh2010advancements,poh2010non} and specific regions of interest (ROI). Furthermore, adaptive temporal filtering techniques, as proposed by Li et al.~\cite{li2014remote}, have also contributed to signal recovery. To enhance the signal-to-noise ratio of recovered rPPG signals, various signal decomposition methods have been introduced, including Independent Component Analysis (ICA)~\cite{poh2010advancements} and matrix completion~\cite{tulyakov2016self}. Moreover, color space projection methods like chrominance subspace\cite{de2013robust} and skin-orthogonal space~\cite{wang2016algorithmic} have been developed to address challenges related to skin tone and head motion. While these approaches have made notable progress in the early stages, they still have limitations. Specifically, they require empirical knowledge to design components, such as hyperparameters in signal processing filtering, and lack supervised learning models that can handle data variations, particularly in demanding environments with significant interference.

\subsection{Supervised}
In recent years, deep learning (DL) techniques for remote photoplethysmography (rPPG) measurement have gained significant traction. Studies \cite{chen2018deepphys,vspetlik2018visual,liu2020multi,nowara2021benefit,li2023learning} have utilized 2D convolutional neural networks (2DCNN) with two consecutive video frames as input to accurately estimate rPPG. Other DL-based approaches \cite{niu2020video,lu2021dual,niu2019rhythmnet,lu2023neuron,du2023dual} have used spatial-temporal signal maps from various facial regions as input for 2DCNN models. Furthermore, 3DCNN-based methods aim to enhance performance, especially for compressed videos \cite{yu2019remote,gideon2021way}. These supervised methods require both facial videos and ground truth physiological signals for training.

Recently, Wang et al.~\cite{wang2022self} proposed a self-supervised rPPG method that does not require ground truth signals during training. However, these studies focus on the estimation of heart rate rather than reconstructing the PPG signal.

\subsection{Unsupervised}
Gideon et al.\cite{gideon2021way} introduced the first unsupervised rPPG method, eliminating the need for ground truth signals during training. Although groundbreaking, this method shows lower accuracy compared to supervised approaches and is more susceptible to external noise. Subsequent unsupervised techniques\cite{sun2022contrast,sun2024contrast,speth2023non,yang2022simper,yue2022video} rely solely on facial videos for training, achieving performance comparable to supervised methods without the need for ground truth signals. This reduces the substantial costs associated with collecting ground truth data. However, unsupervised learning in rPPG faces challenges with bias and distinguishing between different PPG signals, often estimating the same heart rate for multiple users despite variations in PPG signal shapes.

\section{Approach}
\begin{figure*}
    \centering
\includegraphics[width=1.0\linewidth]{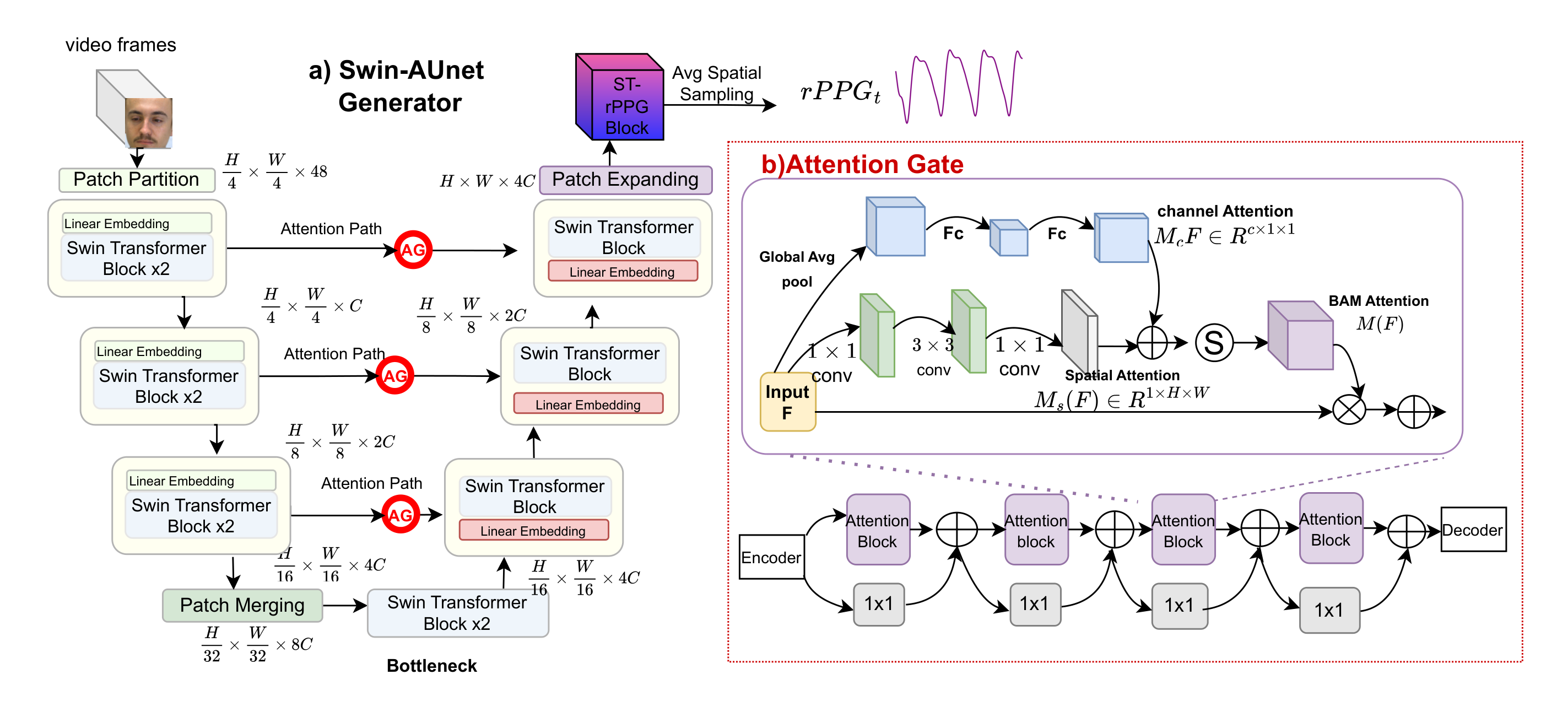}   
    \caption{Network structure: a) Generator: Unet with incorporating Attention Gate and Swin-transformer-V2. b) Attention Gate: capturing temporal dependencies and focusing on relevant facial regions to capture rPPG signal, c) Swin transformer(V2} with additional MLP layer.
    \label{fig:model}

\end{figure*}

\subsection{Generator: Swin-AttentionUnet (Swin-AUnet)}
The proposed generator combines elements from U-Net, the Swin Transformer, and an attention gate mechanism to reconstruct 2D rPPG signals from 3D facial video inputs. This architecture is motivated by U-Net's success in capturing spatial and temporal dependencies, which is essential for accurately modeling rPPG signals. However, adapting U-Net for this task required several modifications to accommodate the input-output dimensional difference.

In our encoder, facial frames are divided into patches and processed through Swin Transformer blocks and patch merging layers to construct hierarchical feature representations. The decoder utilizes a patch-expanding layer for upsampling deep features and a Swin Transformer block to refine features and capture spatial dependencies. The attention gate mechanism replaces traditional skip connections, focusing on relevant facial regions and capturing temporal dependencies to aid rPPG signal extraction.

To enhance feature learning, we incorporate Swin Transformer V2, which employs scaled cosine attention to compute attention logits between pixel pairs, along with additional modifications like an extra MLP layer.

While the current architecture shows promise, we recognize the need for exploring alternative models tailored specifically for rPPG signal reconstruction from facial videos. Future iterations may involve investigating specialized CNN architectures or transformer-based models to better suit the requirements of this task. The choice of architecture will be carefully evaluated based on its ability to model the input-output relationship effectively and capture relevant spatial and temporal information. (See Figures \ref{fig:model} and Figure \ref{fig:architect} for visual representations of the architecture.)

The scaled cosine attention mechanism in Swin Transformer V2 is defined by the equation:

\[
\text{Sim}(q_i, k_j) = \frac{\cos(q_i, k_j)}{\tau} + B_{ij}
\]

Here, \( B_{ij} \) represents the positional relationship bias between pixels \( i \) and \( j \), and \( \tau \) is a trainable scalar unique for each head and layer, not shared between them \cite{liu2022swin}.

By integrating the Swin Transformer, the attention gate, and the architectural modifications, our Swin-AUnet generator aims to effectively reconstruct the rPPG signal from facial videos, capturing the intricate morphological features of the ground-truth PPG waveform \cite{huang2020unet,liu2021swin,liu2022swin}.

\textbf{Regarding the motivation behind using a U-Net architecture, we clarify that while the input is a 3D signal (H × W × T) and the output is a 2D signal (Amplitude × T), we apply spatial global average pooling after the final layer of the U-Net to extract the 2D rPPG signal. This adaptation allows the network to effectively process the 3D input signal and generate the desired 2D output, leveraging U-Net's strengths in capturing both spatial and temporal dependencies.}

\subsection{Multi-Discriminators}
To preserve the time and frequency information on the generated rPPG signal, we use PatchGAN discriminators. Also, to generate more realistic rPPG signal we have a discriminator to distinguish between the second derivative of PPG signal (SDPPG) and generated rPPG signal (Figure~\ref{fig:architect}). The purpose of SDPPG is to capture the onset, systolic and diastolic peaks of PPG signals (See Figure~\ref{fig:wavelet}-left). The frequency discriminator will evaluate frequency spectra of PPG signal and reconstruct (rPPG) such as heart rate and respiratory rate and frequency spectra of systolic, notch, and diastolic peaks.

The motivation behind using multiple discriminators comes from the need to capture different aspects of the rPPG signal, ensuring that the generated waveform successfully mimics the ground-truth morphology PPG signal. By using discriminators in the time domain, frequency domain, and second derivative of the time domain, we can effectively evaluate the quality of the reconstructed signal from various perspectives. This multidiscriminator approach helps to enforce consistency and fidelity across different signal representations, leading to a more accurate and robust estimation of the rPPG.
\subsubsection{Time Domain}
The PatchGAN discriminator is designed to work on sequential data (time series signals) and is trained to distinguish between real PPG and generated rPPG patches ($D_{t}$ in Figure~\ref{fig:architect}). The time domain discriminator is capable of handling the temporal nature of PPG signals.We use recurrent layers or 1D convolution layers to capture the temporal dependencies within the signal.
\subsubsection{Frequency Domain}
 The wavelet transform decomposes the PPG signal into various scales using a single mathematical operation. This allows us to examine both high-frequency and low-frequency components simultaneously, providing a richer understanding of the signal's behavior. The optimal mother wavelet function selection depends on the morphology of PPG signal. In this study, the DB4 wavelet transform demonstrated superior performance because it closely resembles the patterns observed in PPG signals. This can be attributed to its ability to decompose the signal into various frequency scales (as illustrated in Fig.~\ref{fig:wavelet}), which effectively captures the key features that differentiate a true PPG signal from a remote one.


\subsubsection{Second Derivative of Time Domain Signal}
\begin{figure}
    \centering
    \includegraphics[width=1.0\linewidth]{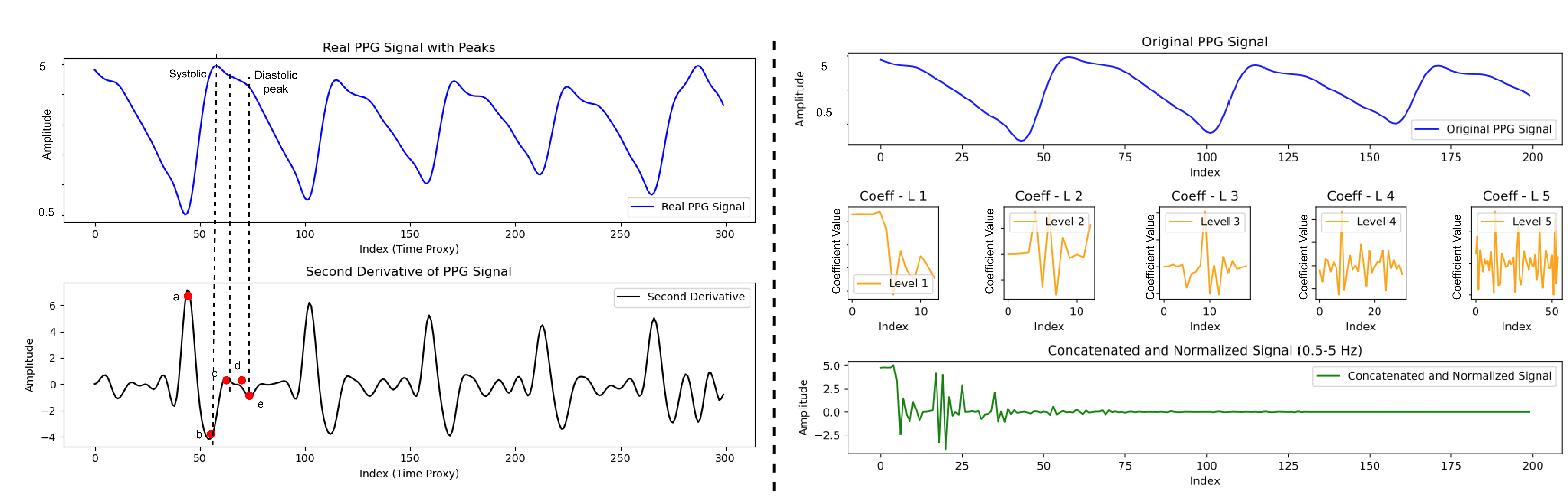}
    \caption{\textbf{Right:} Wavelet analysis of PPG signal reveals frequency components (0.5-5 Hz) through decomposition and reconstruction. \textbf{Left:} Second Derivative of PPG (SDPPG) displays distinct systolic (a, b, c, d) and diastolic (e) waves compared to the original PPG waveform.}
    \label{fig:wavelet}

\end{figure}

Previous research on remote photoplethysmography (rPPG) has primarily focused on heart rate extraction, which may not accurately reflect the morphology of the ground truth PPG signal. Heart rate alone offers limited information about the cardiovascular system and does not provide insight into blood vessel conditions, heart rhythm irregularities, or other detailed hemodynamic parameters. In contrast, PPG morphology provides detailed information about the cardiovascular system, including pulse wave velocity, arterial stiffness, and peripheral vascular disease. By analyzing the shape and features of the PPG waveform, it is possible to detect arrhythmias, heart rate variability, and other cardiac events that are not discernible by heart rate alone. This study is the first to attempt to extract not only the heart rate but also the morphology of the PPG signal using a second derivative loss function, which can effectively mimic the systolic and diastolic phases of the PPG waveform.
Derivatives play a crucial role in enhancing the examination of PPG signals, particularly in peak identification. The first derivative of the PPG signal indicates important events, such as systolic and diastolic peaks~\cite{liu2017cuffless} (Figure~\ref{fig:wavelet}-right). Although the second derivative of the PPG (SDPPG) may exhibit different values and peak positions compared to the original PPG signal, understanding and interpreting these changes are essential for gaining insights into distinctive PPG features~\cite{md2020heart}. Additionally, the PPG waveform comprises one systolic wave and one diastolic wave, whereas the second derivative waveform includes four systolic waves (a, b, c, and d waves) and one diastolic wave (e wave)~\cite{md2020heart} (See Figure~\ref{fig:wavelet}-left).

 The PatchGAN architecture, used for all discriminators, is a modification of the standard discriminator architecture, providing fine-grained feedback at the patch level~\cite{chen2023multi}(Figure~\ref{fig:architect}). Operating at a local patch level, the PatchGAN discriminator uses segmented PPG windows as input in the time domain, frequency domain, and second derivative of the time domain signal instead of making a single decision for the entire signal~\cite{chen2023multi}.
\subsection{Loss Function}
\label{subsection:loss}
The discriminator improves performance by integrating four loss functions: variance loss for noise reduction, dynamic time warping (DTW) loss for alignment, sparsity loss for heart rate adjustment, and variance loss for maintaining a consistent distribution and time interval in the PPG signal frequency domain.

\subsubsection{Sparsity Loss}
Sparsity Loss improves heart rate precision by considering time, frequency, and the second derivative of the signal in the time domain~\cite{speth2023non}. It quantifies the sum of absolute amplitudes of discrete time signals ($x[n]$) within specified intervals:
\begin{equation}
L_{s_{time}} = \sum_{n=a}^{N}\left | x[n] \right |
\end{equation}
\begin{equation}
L_{s_{sd}} = \sum_{n=a}^{N}\left | x''[n] \right |
\end{equation}
In the frequency domain, Sparsity Loss ($L_{s_{freq}}$) is a weighted sum around the spectral peak frequency:
\begin{equation}
L_{s_{freq}} = \frac{\sum_{i=a}^{\text{argmax}(F)-\Delta F}F_{i} + \sum_{\text{argmax}(F)+\Delta F}^{b}F_{i}} {\sum_{i=a}^{b}F_{i}}
\end{equation}
These losses minimize non-significant amplitudes, highlight specific frequency components, and emphasize sharp changes in the second derivative, guiding the model to prioritize crucial periodic components associated with the pulse.

\subsubsection{Variance Loss}
Variance loss distinguishes between generated rPPG and actual PPG signals in the time domain (original signal and second derivative) and in the frequency domain (wavelet db4)~\cite{speth2023non}. The loss function is:
\begin{equation}
L_{\text{domain}} = \frac{1}{d}\sum_{i=1}^{d}(CDF_{i}(Q_{\text{domain}})-CDF_{i}(P_{\text{domain}}))^2
\end{equation}
where \( d \) is the number of frequencies, and \( CDF_{i}(Q_{\text{domain}}) \) and \( CDF_{i}(P_{\text{domain}}) \) are the Cumulative Distribution Functions of the predicted rPPG and ground truth PPG, respectively. This loss captures distributional differences and maintains uniformity across the desired frequency domain and time intervals.

\subsubsection{Soft Dynamic Time Warping Loss (Soft-DTW)}
Soft Dynamic Time Warping (SoftDTW)~\cite{cuturi2017soft} is a differentiable loss for comparing time-series data. It measures the similarity between PPG $(x_{1},...,x_{n})\in \mathbb{R}^{p\times n}$ and rPPG $(y_{1},...,y_{m})\in \mathbb{R}^{p\times m}$:
\begin{equation}
DTW(x,y) := \min_{A \in A_{n,m}}\left \langle A,\Delta (x,y) \right \rangle
\end{equation}
where $\Delta(x,y) = \left [ \delta (x_{i}, y_{i}) \right ]_{i,j} \in \mathbb{R}^{n\times m}$ is the cost matrix. The soft-minimum operator is defined as:
\begin{equation}
\min_{\gamma} (a_{1},...,a_{n})=\left\{\begin{matrix}
\min_{i} a_{i} & \text{if} \ \gamma = 0 \\ 
-\gamma \log \sum_{i} e^{-a_{i}/\gamma} & \text{if} \ \gamma > 0
\end{matrix}\right.
\end{equation}
leading to:
\begin{equation}
dtw_{\gamma}(x,y) = \left ( \frac{\partial \Delta (x,y)}{\partial x} \right )^ T A^* 
\end{equation}
and for $\gamma > 0$:
\begin{equation}
\bigtriangledown _xdtw_{\gamma}(x,y) = \left ( \frac{\partial \Delta (x,y)}{\partial x} \right )^ T E_\gamma\left [ A \right ]
\end{equation}
The final loss function is $Loss(x,y) = dtw_{\gamma}(x,y)$.

\subsubsection{Final Loss}
The combined loss function $L_{\text{total}}$ is a weighted sum of individual losses:
\begin{align*}
L_{\text{total}} &= \alpha(L_{\text{time}}) + \beta(L_{\text{freq}}) + \gamma(L_{\text{sd}}) \\
&= \alpha(L_{DTW(t)} + L_{sparsity(t)} + L_{variance(t)}) \\
& + \beta(L_{sparsity(freq)} + L_{variance(freq)}) \\
& +\gamma(L_{DTW(sd)} + L_{sparsity(sd)} + L_{variance(sd)}) \\
\end{align*}
where $\alpha$, $\beta$, and $\gamma$ are loss coefficients for $D_{time}$, $D_{freq}$, and $D_{sd}$. Weighting these losses improves performance by guiding the model to explore frequencies and spatial time within video data, aiming for better representation features resembling real PPG signals. Incorporating these complementary loss functions significantly enhances the model's ability to reproduce intricate PPG signal features, including systolic and diastolic components.

During training, we adjusted the weights of the loss functions to enhance the performance of the model. We prioritized accurate PPG waveform reconstruction by increasing the time-domain loss weight (\(\alpha = 1.5\)) and slightly reducing the frequency-domain weight (\(\beta = 0.8\)). Additionally, we raised the second derivative loss weight (\(\gamma = 1.2\)) to improve the detection of systolic and diastolic peaks. These adjustments were guided by validation results and qualitative waveform analysis.

\section{Experiments}
\subsection{Dataset}
In our project, we conducted an assessment using five well-established remote photoplethysmography (rPPG) datasets, encompassing both RGB and near-infrared (NIR) videos captured in diverse scenarios. The datasets selected for analysis are PURE~\cite{stricker2014non}, UBFC-rPPG~\cite{bobbia2019unsupervised}, OBF~\cite{li2018obf}, MR-NIRP~\cite{magdalena2018sparseppg}, and MMSE-HR~\cite{zhang2016multimodal}. 

\begin{itemize}
    \item PURE: It comprises video recordings of 10 subjects, each captured in 6 sessions lasting approximately 1 minute. The raw video data was collected at 30 frames per second (fps).
    \item UBFC-rPPG: This data set consists of 1-minute video recordings from 42 subjects captured at 30 frames per second (fps). To ensure data quality, head movement was minimized while subjects participated in a time-sensitive math game designed to elevate their heart rates
    \item OBF: This dataset consists of videos from 100 healthy subjects recorded both before and after exercises. 
    \item MR-NIRP: This dataset encompasses NIR videos from eight subjects in stationary or motion tasks~\cite{nowara2020near,magdalena2018sparseppg}. 
    \item MMSE-HR: This data set, comprising 102 videos from 40 subjects recorded during emotion elicitation experiments was used. This data set presents challenges due to the participation of spontaneous facial expressions and head motions~\cite{zhang2016multimodal}.
\end{itemize}

\subsection{Preprocessing}
The process starts by pre-processing original videos to extract the facial region. Facial landmarks are generated using OpenFace~\cite{baltrusaitis2018openface}, and a boundary box is defined based on these landmarks. This bounding box, which encompasses facial features, is applied to crop the face from the video. The resulting face crops are resized to $128 \times 128$ pixels as inputs from the Swin-AUnet generator.
\subsection{Evaluation Metrics}
Similar to previous studies \cite{yu2019remote,niu2020video,song2021pulsegan,lu2021dual}, we evaluated our model's performance by comparing its estimated heart rate (HR) with ground truth data from various datasets. The HR evaluation metrics included mean absolute error (MAE), root mean squared error (RMSE) and Pearson's correlation coefficient (R), following the approaches in~\cite{song2021pulsegan,gideon2021way}.

In addition to these metrics for HR, we assessed the quality of the reconstructed rPPG signals by comparing them to the ground-truth PPG signal using several quantitative metrics. This comprehensive approach provides a fuller picture of the reconstructed rPPG signal's accuracy, beyond methods focusing solely on HRV. Pearson's correlation coefficient ($\rho$) measures how closely PPG signals and their reconstructed versions (rPPG) correlate, ranging from 0 (weak correlation) to 1 (strong correlation).

\begin{equation}
\rho=\frac{\left(y_{PPG}-\bar{y}{PPG}\right)^T\left(y_{rPPG}-\bar{y}_{rPPG}\right)}{\left|y_{PPG}-\bar{y}_{PPG}\right|^2\left|y_{rPPG}-\bar{y}_{rPPG}\right|^2}
\end{equation}

where $y_{PPG}$ and $y_{rPPG}$ represent the original PPG signal and its reconstructed version, respectively. The function $||*||_2$ denotes the Euclidean distance.

The Fréchet distance (FD) assesses the similarity between two curves, considering both the location and the order of points. It compares the original PPG waveform (ground truth) with its rPPG signal.

\begin{equation}
F D=\min \left(\max {i \in Q}\left(d\left(y_{PPG_i}, y_{rPPG_i}\right)\right)\right), Q=[1, m]
\end{equation}

FD relies on Euclidean distance ($d(*)$) to measure the similarity between ground truth PPG and rPPG waveforms. Additionally, RMSE measures the average magnitude of the differences between the ground truth PPG signal and its reconstructed rPPG signal. \textit{Our work distinguishes itself from previous research by directly evaluating the similarity between the reconstructed rPPG signal and the ground truth PPG signal. Previous approaches primarily focused on heart rate extraction, not the fidelity of the reconstructed signal itself.}
\begin{figure*}
    \centering
    \includegraphics[width=1.0\linewidth]{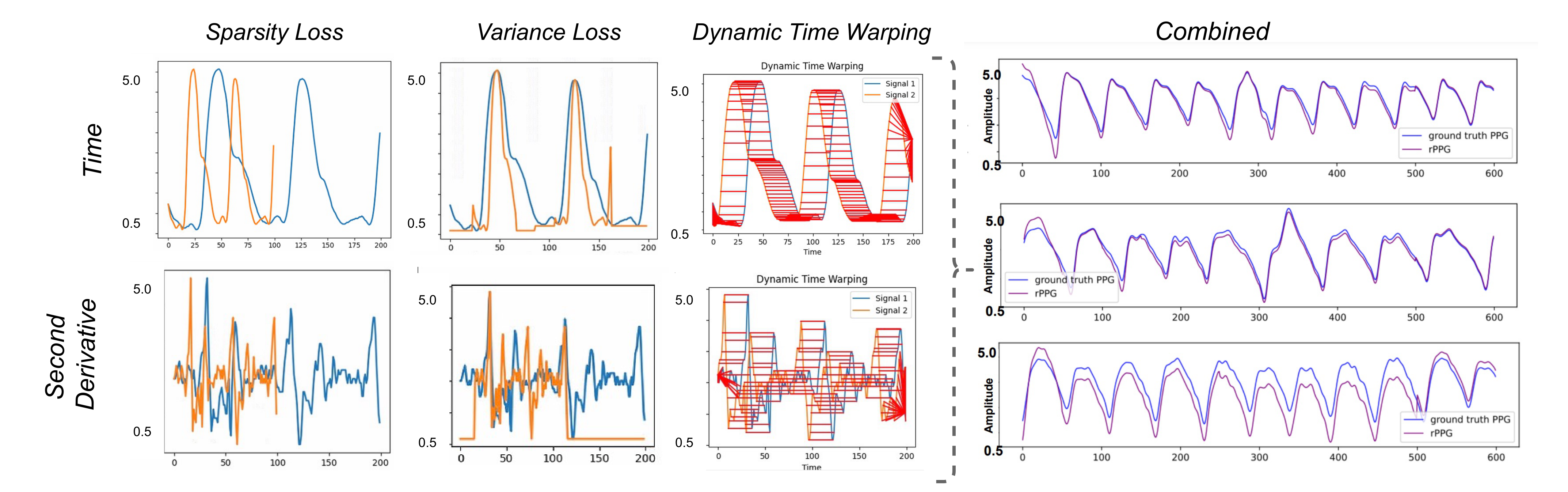}    
\caption{Comparison of different loss functions (Sparsity Loss, Variance Loss, Dynamic Time Warping, and Combined) across various time derivatives (0.5 and 5.0 seconds). The table shows the performance of each loss function in capturing temporal patterns and handling time series data.}
    \label{fig:PPG-rPPG}

\end{figure*}
\subsection{Implementation Details}
We evaluated our model using within-dataset and cross-dataset testing. For within-dataset, we split each dataset into training (first 40 seconds of video) and testing (remaining video). For cross-dataset, we trained on one entire dataset and tested on a different dataset. The Swin-AUnet model was trained on 2x NVIDIA RTX 4090 GPUs, using the Adam optimizer with a 0.001 learning rate over 400 epochs.

\subsection{Results}
We have trained our Swin-AUnet architecture with a multidiscriminator on five different datasets and evaluated the performance. According to Table~\ref{tab:results}, our model has a significant improvement in measuring HR across different datasets. Swin-AUnet imporve HR measurement in UBFC by 18\% than Dual-GAN~\cite{lu2021dual}, 71. 95\% in the PURE dataset than Dual-GAN~\cite{lu2021dual}, 3. 92\% in the OBF dataset than Contrast-phys~\cite{sun2022contrast} and Contrast-phys+, 61.96\% in MR-NIRP than Pulse-GAN~\cite{song2021pulsegan}, and 27.04\% in MMSE-HR than the PhysNet model~\cite{yu2019remote}.
For the cross-dataset testing scenario, we train on UBFC-rPPG and test on the MMSE-HR dataset(Table~\ref{tab:cross}). Also, for intra-cross validation, we train and test on the MMSE-HR dataset (Table~\ref{tab:cross}). Our results show that the Swin-AUnet successfully improves performance on both cross-dataset and intra-dataset. The MAE improved by 44.31\% than Contrast-phys+~\cite{sun2024contrast} in cross-dataset validation for UBFC and MMSE-HR datasets, and 79.27\% than Contrast-phys+~\cite{sun2024contrast} in intra-dataset validation for MMSE-HR dataset.
In Figure~\ref{fig:PPG-rPPG}, we compare the ground truth PPG signal against the rPPG signal generated using Swin-AUnet.
\begin{table*}[ht]
\begin{center}
\resizebox{\textwidth}{!}{%
\begin{tabular}{|c|ccc|ccc|ccc|ccc|ccc|}
\hline
\multirow{2}{*}{Methood} &
  \multicolumn{3}{c|}{UBFC-rPPG} &
  \multicolumn{3}{c|}{PURE} &
  \multicolumn{3}{c|}{OBF} &
  \multicolumn{3}{c|}{MR-NIRP (NIR)}&
  \multicolumn{3}{c|}{MMSE-HR}\\ \cline{2-16} 
 &
  \multicolumn{1}{c|}{\begin{tabular}[c]{@{}c@{}}MAE\\ (bpm)\end{tabular}} &
  \multicolumn{1}{c|}{\begin{tabular}[c]{@{}c@{}}RMSE\\ (bpm)\end{tabular}} &
  R &
  \multicolumn{1}{c|}{\begin{tabular}[c]{@{}c@{}}MAE\\ (bpm)\end{tabular}} &
  \multicolumn{1}{c|}{\begin{tabular}[c]{@{}c@{}}RMSE\\ (bpm)\end{tabular}} &
  R &
  \multicolumn{1}{c|}{\begin{tabular}[c]{@{}c@{}}MAE\\ (bpm)\end{tabular}} &
  \multicolumn{1}{c|}{\begin{tabular}[c]{@{}c@{}}RMSE\\ (bpm)\end{tabular}} &
  R &
  \multicolumn{1}{c|}{\begin{tabular}[c]{@{}c@{}}MAE\\ (bpm)\end{tabular}} &
  \multicolumn{1}{c|}{\begin{tabular}[c]{@{}c@{}}RMSE\\ (bpm)\end{tabular}} &
  R &
    \multicolumn{1}{c|}{\begin{tabular}[c]{@{}c@{}}MAE\\ (bpm)\end{tabular}} &
  \multicolumn{1}{c|}{\begin{tabular}[c]{@{}c@{}}RMSE\\ (bpm)\end{tabular}} &
  R \\ \hline
GREEN~\cite{verkruysse2008remote} &
  \multicolumn{1}{c|}{7.5} &
  \multicolumn{1}{c|}{14.41} &
  0.62 &
  \multicolumn{1}{c|}{-} &
  \multicolumn{1}{c|}{-} &
  - &
  \multicolumn{1}{c|}{-} &
  \multicolumn{1}{c|}{2.162} &
  0.99 &
  \multicolumn{1}{c|}{-} &
  \multicolumn{1}{c|}{-} &
  - &\multicolumn{1}{c|}{-} &
  \multicolumn{1}{c|}{-} &
  -2 \\ \hline
ICA~\cite{poh2010advancements} &
  \multicolumn{1}{c|}{5.17} &
  \multicolumn{1}{c|}{11.76} &
  0.65 &
  \multicolumn{1}{c|}{-} &
  \multicolumn{1}{c|}{-} &
  - &
  \multicolumn{1}{c|}{-} &
  \multicolumn{1}{c|}{2.73} &
  0.98 &
  \multicolumn{1}{c|}{-} &
  \multicolumn{1}{c|}{-} &
  - &\multicolumn{1}{c|}{-} &
  \multicolumn{1}{c|}{-} &
  - \\ \hline
CHROM~\cite{de2013robust} &
  \multicolumn{1}{c|}{2.37} &
  \multicolumn{1}{c|}{4.91} &
  0.89 &
  \multicolumn{1}{c|}{2.07} &
  \multicolumn{1}{c|}{9.92} &
  0.99 &
  \multicolumn{1}{c|}{-} &
  \multicolumn{1}{c|}{2.733} &
  0.98 &
  \multicolumn{1}{c|}{-} &
  \multicolumn{1}{c|}{-} &
  - &\multicolumn{1}{c|}{-} &
  \multicolumn{1}{c|}{13.97} &
  0.55 \\ \hline
2SR~\cite{wang2015novel} &
  \multicolumn{1}{c|}{-} &
  \multicolumn{1}{c|}{-} &
  - &
  \multicolumn{1}{c|}{2.44} &
  \multicolumn{1}{c|}{3.06} &
  0.98 &
  \multicolumn{1}{c|}{-} &
  \multicolumn{1}{c|}{-} &
  - &
  \multicolumn{1}{c|}{-} &
  \multicolumn{1}{c|}{-} &
  - &\multicolumn{1}{c|}{-} &
  \multicolumn{1}{c|}{-} &
  - \\ \hline
POS~\cite{wang2016algorithmic} &
  \multicolumn{1}{c|}{4.05} &
  \multicolumn{1}{c|}{8.75} &
  0.78 &
  \multicolumn{1}{c|}{-} &
  \multicolumn{1}{c|}{-} &
  - &
  \multicolumn{1}{c|}{-} &
  \multicolumn{1}{c|}{1.906} &
  0.991 &
  \multicolumn{1}{c|}{-} &
  \multicolumn{1}{c|}{-} &
  - &\multicolumn{1}{c|}{-} &
  \multicolumn{1}{c|}{-} &
  - \\ \hline
Meta-rPPG~\cite{lee2020meta} &
  \multicolumn{1}{c|}{5.97} &
  \multicolumn{1}{c|}{7.42} &
  0.53 &
  \multicolumn{1}{c|}{-} &
  \multicolumn{1}{c|}{-} &
  - &
  \multicolumn{1}{c|}{-} &
  \multicolumn{1}{c|}{-} &
  - &
  \multicolumn{1}{c|}{-} &
  \multicolumn{1}{c|}{-} &
  - &\multicolumn{1}{c|}{-} &
  \multicolumn{1}{c|}{-} &
  - \\ \hline
Dual-GAN~\cite{lu2021dual} &
  \multicolumn{1}{c|}{0.44} &
  \multicolumn{1}{c|}{0.67} &
  0.99 &
  \multicolumn{1}{c|}{0.82} &
  \multicolumn{1}{c|}{1.31} &
  0.99 &
  \multicolumn{1}{c|}{-} &
  \multicolumn{1}{c|}{-} &
  - &
  \multicolumn{1}{c|}{-} &
  \multicolumn{1}{c|}{-} &
  - &\multicolumn{1}{c|}{-} &
  \multicolumn{1}{c|}{-} &
  - \\ \hline
PhysNet~\cite{yu2019remote} &
  \multicolumn{1}{c|}{-} &
  \multicolumn{1}{c|}{-} &
  - &
  \multicolumn{1}{c|}{2.1} &
  \multicolumn{1}{c|}{2.6} &
  0.99 &
  \multicolumn{1}{c|}{-} &
  \multicolumn{1}{c|}{1.812} &
  0.992 &
  \multicolumn{1}{c|}{3.07} &
  \multicolumn{1}{c|}{7.55} &
  0.655 &\multicolumn{1}{c|}{1.22} &
  \multicolumn{1}{c|}{4.49} &
  0.94 \\ \hline
rPPGNet~\cite{yu2019remote} &
  \multicolumn{1}{c|}{-} &
  \multicolumn{1}{c|}{-} &
  - &
  \multicolumn{1}{c|}{-} &
  \multicolumn{1}{c|}{-} &
  - &
  \multicolumn{1}{c|}{-} &
  \multicolumn{1}{c|}{1.8} &
  0.992 &
  \multicolumn{1}{c|}{-} &
  \multicolumn{1}{c|}{-} &
  - &\multicolumn{1}{c|}{-} &
  \multicolumn{1}{c|}{-} &
  - \\ \hline
Contrast-Phys~\cite{sun2022contrast}&
  \multicolumn{1}{c|}{0.64} &
  \multicolumn{1}{c|}{1.00} &
  0.99 &
  \multicolumn{1}{c|}{1.00} &
  \multicolumn{1}{c|}{1.4} &
  0.99 &
  \multicolumn{1}{c|}{0.51} &
  \multicolumn{1}{c|}{1.39} &
  0.994 &
  \multicolumn{1}{c|}{2.68} &
  \multicolumn{1}{c|}{4.77} &
  0.85 &\multicolumn{1}{c|}{-} &
  \multicolumn{1}{c|}{-} &
  - \\ \hline
PulseGAN~\cite{song2021pulsegan} &
  \multicolumn{1}{c|}{1.19} &
  \multicolumn{1}{c|}{2.10} &
  0.98 &
  \multicolumn{1}{c|}{-} &
  \multicolumn{1}{c|}{-} &
  - &
  \multicolumn{1}{c|}{-} &
  \multicolumn{1}{c|}{-} &
  - &
  \multicolumn{1}{c|}{-} &
  \multicolumn{1}{c|}{-} &
  - &\multicolumn{1}{c|}{7.5} &
  \multicolumn{1}{c|}{14.41} &
  0.62 \\ \hline
Nowara2021~\cite{nowara2021benefit} &
  \multicolumn{1}{c|}{-} &
  \multicolumn{1}{c|}{-} &
  - &
  \multicolumn{1}{c|}{-} &
  \multicolumn{1}{c|}{-} &
  - &
  \multicolumn{1}{c|}{-} &
  \multicolumn{1}{c|}{-} &
  - &
  \multicolumn{1}{c|}{2.34} &
  \multicolumn{1}{c|}{4.46} &
  0.85&\multicolumn{1}{c|}{-} &
  \multicolumn{1}{c|}{-} &
 - \\ \hline
Gideon2021~\cite{gideon2021way} &
  \multicolumn{1}{c|}{1.85} &
  \multicolumn{1}{c|}{4.28} &
  0.93 &
  \multicolumn{1}{c|}{2.3} &
  \multicolumn{1}{c|}{2.9} &
  0.99 &
  \multicolumn{1}{c|}{2.83} &
  \multicolumn{1}{c|}{7.88} &
  0.825 &
  \multicolumn{1}{c|}{4.75} &
  \multicolumn{1}{c|}{9.14} &
  0.61 &\multicolumn{1}{c|}{3.98} &
  \multicolumn{1}{c|}{9.65} &
  0.85 \\ \hline
Contrast-phys+~\cite{sun2024contrast} &
  \multicolumn{1}{c|}{0.64} &
  \multicolumn{1}{c|}{1.00} &
  0.99 &
  \multicolumn{1}{c|}{1.00} &
  \multicolumn{1}{c|}{1.40} &
  0.99 &
  \multicolumn{1}{c|}{0.51} &
  \multicolumn{1}{c|}{1.39} &
  0.994 &
  \multicolumn{1}{c|}{2.68} &
  \multicolumn{1}{c|}{4.77} &
  0.85 &\multicolumn{1}{c|}{3.83} &
  \multicolumn{1}{c|}{0.96} &
  3.72 \\ \hline
PhysFormer~\cite{yu2022physformer} &
  \multicolumn{1}{c|}{-} &
  \multicolumn{1}{c|}{-} &
  - &
  \multicolumn{1}{c|}{-} &
  \multicolumn{1}{c|}{-} &
  - &
  \multicolumn{1}{c|}{-} &
  \multicolumn{1}{c|}{0.804} &
  0.998 &
  \multicolumn{1}{c|}{-} &
  \multicolumn{1}{c|}{-} &
  - &\multicolumn{1}{c|}{1.48} &
  \multicolumn{1}{c|}{4.22} &
 0.95 \\ \hline
\textbf{Swin-AUnet(Ours)} &
  \multicolumn{1}{c|}{\textbf{0.36$\downarrow$}} &
  \multicolumn{1}{c|}{\textbf{0.51$\downarrow$}} &
  \textbf{0.99$\uparrow$} &
  \multicolumn{1}{c|}{\textbf{0.23$\downarrow$}} &
  \multicolumn{1}{c|}{\textbf{0.36$\downarrow$}} &
  \textbf{0.99$\uparrow$} &
  \multicolumn{1}{c|}{\textbf{0.49$\downarrow$}} &
  \multicolumn{1}{c|}{\textbf{0.51$\downarrow$}} &
  \textbf{0.99$\uparrow$} &
  \multicolumn{1}{c|}{\textbf{0.89$\downarrow$}} &
  \multicolumn{1}{c|}{\textbf{0.74$\downarrow$}} &
  \textbf{0.99$\uparrow$} &\multicolumn{1}{c|}{{\textbf{0.89$\downarrow$}}} &
  \multicolumn{1}{c|}{{\textbf{0.75$\downarrow$}}} &
  {\textbf{0.99$\uparrow$}} \\ \hline
\end{tabular}%
}
\caption{HR results. The best results are in bold.}
\label{tab:results}
\end{center}

\end{table*}
Notably, our model adeptly produces an rPPG signal that precisely reconstructs the morphology features of the PPG signal, encompassing both systolic and diastolic phases. As
\begin{table*}[htb]
\begin{center}
\resizebox{\textwidth}{!}{%
\begin{tabular}{|l|ccc|ccc|}
\hline
\multirow{2}{*}{Method} &
  \multicolumn{3}{c|}{Cross-dataset(UBFC $\rightarrow$ MMSE-HR)} &
  \multicolumn{3}{c|}{Intra-dataset(MMSE-HR$\rightarrow$ MMSE-HR)} \\ \cline{2-7} 
 &
  \multicolumn{1}{c|}{MAE(bpm)} &
  \multicolumn{1}{c|}{RMSE(bpm)} &
  \multicolumn{1}{c|}{R} &
  \multicolumn{1}{c|}{MAE(bpm)} &
  \multicolumn{1}{c|}{RMSE(bpm)} &
  \multicolumn{1}{c|}{R}  \\ \hline
CHROM~\cite{de2013robust} &
  \multicolumn{1}{c|}{-} &
  \multicolumn{1}{c|}{13.97} &
  \multicolumn{1}{c|}{0.55} &
  \multicolumn{1}{c|}{-} &
  \multicolumn{1}{c|}{13.97} &
  \multicolumn{1}{c|}{0.55} 
 \\ \hline
PhysNet~\cite{yu2019remote} &
  \multicolumn{1}{c|}{2.04} &
  \multicolumn{1}{c|}{6.85} &
  \multicolumn{1}{c|}{0.86} &
  \multicolumn{1}{c|}{1.22} &
  \multicolumn{1}{c|}{4.49} &
  \multicolumn{1}{c|}{0.94} 
 \\ \hline
PhysFormer~\cite{yu2022physformer} &
  \multicolumn{1}{c|}{2.68} &
  \multicolumn{1}{c|}{7.01} &
  \multicolumn{1}{c|}{0.86} &
  \multicolumn{1}{c|}{1.48} &
  \multicolumn{1}{c|}{4.22} &
  \multicolumn{1}{c|}{0.95} 
 \\ \hline
\begin{tabular}[c]{@{}l@{}}Contrast-Phys~\cite{sun2024contrast}\\ (supervised)\end{tabular} &
  \multicolumn{1}{c|}{1.76} &
  \multicolumn{1}{c|}{5.34} &
  \multicolumn{1}{c|}{0.92} &
  \multicolumn{1}{c|}{1.11} &
  \multicolumn{1}{c|}{3.83} &
  \multicolumn{1}{c|}{0.96} 
  \\ \hline
\begin{tabular}[c]{@{}l@{}}Contrast-Phys~\cite{sun2024contrast}\\ (semi-spervised)\end{tabular} &
  \multicolumn{1}{c|}{2.30} &
  \multicolumn{1}{c|}{6.32} &
  \multicolumn{1}{c|}{0.89} & 
  \multicolumn{1}{c|}{1.20} &
  \multicolumn{1}{c|}{3.89} &
  \multicolumn{1}{c|}{0.96} 
   \\ \hline
\begin{tabular}[c]{@{}l@{}}Contrast-Phys~\cite{sun2022contrast}\\ (unsupervised)\end{tabular} &
  \multicolumn{1}{c|}{2.43} &
  \multicolumn{1}{c|}{7.34} &
  \multicolumn{1}{c|}{0.86} &
  \multicolumn{1}{c|}{1.82} &
  \multicolumn{1}{c|}{6.69} &
  \multicolumn{1}{c|}{0.87} 
 \\ \hline
Gideon2021~\cite{gideon2021way} &
  \multicolumn{1}{c|}{4.10} &
  \multicolumn{1}{c|}{11.55} &
  \multicolumn{1}{c|}{0.70} &
  \multicolumn{1}{c|}{3.98} &
  \multicolumn{1}{c|}{9.65} &
  \multicolumn{1}{c|}{0.85} 
   \\ \hline
\textbf{Swin-AUnet(ours)} &
  \multicolumn{1}{c|}{\textbf{0.98$\downarrow$}} &
  \multicolumn{1}{c|}{\textbf{1.16$\downarrow$}} &
  \multicolumn{1}{c|}{\textbf{0.96$\uparrow$}} &
  \multicolumn{1}{c|}{\textbf{0.23$\downarrow$}} &
  \multicolumn{1}{c|}{\textbf{0.95$\downarrow$}} &
  \multicolumn{1}{c|}{\textbf{0.99$\uparrow$}}
 \\ \hline
\end{tabular}%
}
\caption{Cross-dataset and intra-dataset HR. The best results are in bold.}
\label{tab:cross}
\end{center}

\end{table*}
can be seen in Table~\ref{tab:similarity}, our proposed Swin-AUnet outperformed compared to Contrast-phys~\cite{sun2022contrast} and Dual-GAN~\cite{lu2021dual}.
Our proposed model achieved a $\rho$ value of $0.915$ indicating a strong correlation between the reconstructed rPPG signal and the PPG signal. Moreover, low values of RMSE (0.167) and FD (0.248) demonstrate the strong similarity between ground truth PPG and rPPG. Moreover, the low RSME and FD support the diversity of the rPPG signal extracted from facial video. Compared to Contrast-phys~\cite{sun2022contrast} and Dual-GAN~\cite{lu2021dual}, our proposed model has improved the mean and standard division by more than 40\%.
\begin{table}[htbp]
\centering
\begin{tabular}{lcccccc}
\hline
\multirow{2}{*}{Models} & \multicolumn{2}{c}{RMSE} & \multicolumn{2}{c}{FD} & \multicolumn{2}{c}{$\rho$} \\
\cline{2-7}
 & $\mu$ & $\sigma$ & $\mu$ & $\sigma$ & $\mu$ & $\sigma$ \\
\hline
\textbf{Swin-AUnet} & $\underline{\mathbf{0.167}}$ & 0.059 & $\underline{\mathbf{0.248}}$ & 0.185 & $\underline{\mathbf{0.915}}$ & 0.057 \\
Contrast-phys~\cite{sun2022contrast} & $\underline{0.469}$ & 0.279 & $\underline{0.538}$ & 0.225 & $\underline{0.621}$ & 0.251 \\
Dual-GAN~\cite{lu2021dual} & $\underline{0.356}$ & 0.346 & $\underline{0.471}$ & 0.118 & $\underline{0.703}$ & 0.117 \\
\hline
\end{tabular}
\caption{Comparison of rPPG signal estimation with previous work - $\mu$ and $\sigma$ are mean and standard deviation, respectively. The evaluation is based on how our proposed model resembles ground truth PPG based on Swin-AUnet compared to other methods. The results are demonstrated based on average across five different datasets.}
\label{tab:similarity}

\end{table}

\subsection{Ablation Study}
\label{subsection:ablation}
To analyze the contribution of each component in our proposed framework, we perform an ablation study evaluating the performance of the model with different combinations of loss functions and architectural modifications. Table~\ref{tab:abl} presents the results of this study on the MMSE-HR data set.
The results in Table~\ref{tab:abl} demonstrate the effectiveness of each loss component in improving the model performance. The variance loss helps distinguish between real and generated signals, leading to a significant reduction in MAE and RMSE. The loss of dynamic time warping (DTW) further enhances the temporal alignment of the predicted rPPG signal, resulting in improved correlation with ground truth. The sparsity loss helps to adjust the heart rate by promoting sparse representations, while the variance loss ensures a consistent distribution across the desired frequency domain and time intervals between the systolic and diastolic phases.
Figure~\ref{fig:PPG-rPPG} presents qualitative visualizations of the predicted rPPG signals with different loss combinations. The baseline model struggles with accurately capturing the waveform shape, but as we add variance, DTW, sparsity, and variance losses, the predicted signal increasingly resembles the ground truth, faithfully reproducing systolic and diastolic peaks.
The ablation study shows the benefits of the architectural modifications of our model. Incorporating the Swin Transformer module helps capture long-range dependencies in the input video, enhancing performance. The attention gate further improves the model's focus on relevant facial regions for rPPG estimation, resulting in additional gains.
While a simple convolutional network could suffice, we found the U-Net architecture, with its skip connections and multi-scale feature fusion, more effective in mapping facial videos to rPPG signals. The U-Net's encoder-decoder structure extracts relevant features at various scales, combining them to produce the rPPG output.
The ablation study also examines the impact of including the second derivative of the time domain signal in the loss functions. Figure~\ref{fig:PPG-rPPG} compares the predicted rPPG signal with and without second derivative losses ($L_{vSD}$, $L_{sSD}$, and $L_{DTW_{SD}}$). The inclusion of these losses results in a smoother and more defined rPPG waveform, especially in the diastolic phase, capturing subtle signal changes and yielding a more accurate representation of systolic and diastolic peaks. 
This improvement is reflected in lower MAE and RMSE values with the complete loss function combination, including the second derivative terms (marked with $\star$ in Table \ref{tab:abl}).
\begin{table}[htb]
\centering

\begin{tabular}{lccc}
\toprule
Loss functions (T,F) & MAE $\downarrow$ & RMSE $\downarrow$ & FD $\uparrow$ \\
\midrule
$L_{vT} + L_{vF}$ & 20.32 &30.5  &  0.547\\
$L_{sT}+L_{sF}$ & 40.03 & 48.63& 0.569 \\
$L_{DTW}$ & 38.2 & 47.32 &  0.426\\
$L_{vT} + L_{vF}+L_{sT}+L_{sF}+L_{DTW}$& 35.32 & 45.2 & 0.462\\
\toprule
Loss functions (T,F,SD) & MAE $\downarrow$ & RMSE $\downarrow$ & FD $\uparrow$ \\
\midrule
$L_{vT} + L_{vF} +L_{vSD}$ & 12.54 & 29.4 & 0.641 \\
$L_{sT}+L_{sF}+ L_{sSD}$ & 38.96& 45.78& 0.531\\
$L_{DTW}+L_{DTW_{SD}}$ & 36.43 & 44.95 &  0.567\\
$L_{vT} + L_{vF} +L_{vSD}+L_{sT}+L_{sF}+L_{sSD}+L_{DTW}+L_{DTW_{SD}}\star$&\textbf{0.89} &\textbf{0.75} & \textbf{0.248} \\
\midrule
Comparison of different architectures with final loss: &MAE $\downarrow$ & RMSE $\downarrow$ & FD $\uparrow$ \\
\midrule
U-Net & 2.54 & 3.03 & 0.302\\
U-Net + ST & 3.61 & 3.02 & 0.298 \\
U-Net + AG & 3.25& 2.94 &0.296 \\
\textbf{U-Net + ST + AG} & \textbf{0.89} & \textbf{0.75} & \textbf{0.248}\\
\bottomrule
\end{tabular}

\caption{Ablation study on the MMSE-HR dataset. $\downarrow$ and $\uparrow$ denote if lower or higher values are preferred. T and F represent the time and frequency domains. $L_{vT}$, $L_{vF}$ are variance losses; $L_{sT}$, $L_{sF}$ are sparsity losses. $L_{DTW}$ is Dynamic Time Warping loss. 'Second derivative' refers to the time domain's second derivative signal. $L_{vSD}$, $L_{sSD}$ are variance and sparsity losses for this derivative. $L_{DTW_{SD}}$ is the DTW loss for this derivative. $\star$ marks the final loss combination in Swin-AUnet.}
\label{tab:abl}

\end{table}

\section{Conclusion}
In this paper, we propose the Swin-AUnet model to reconstruct the remote photoplethysmography (rPPG) based on facial video recorded from five datasets. The proposed Swin-AUnet model comprises Unet architecture with a Swin transformer module replaced by CNN and an attention gate which was replaced by a skip connection. To train the mode, we used patch GAN with multiple discriminators to preserve the time and frequency information on the generated rPPG signal. The discriminators will evaluate the quality of rPPG signal by comparing the time domain (original signal and second derivative), frequency domain (wavelet db4) to resemble the morphology of the ground truth PPG signal. The discriminator enhances performance through the integration of four loss functions. variance loss addresses noise challenges, dynamic time warping loss manages alignment issues, sparsity loss adjusts heart rate, and variance loss maintains a consistent distribution and time interval in the PPG signal’s frequency domain. We have finally evaluated the proposed model under different scenarios such as within-dataset testing and cross-dataset testing. Our experimental results demonstrated support for the effectiveness of the proposed model and showed improvement in HR estimations and also waveform reconstruction. Our work distinguishes itself from prior research by directly evaluating the similarity between the reconstructed rPPG signal and the ground truth PPG signal. Previous approaches primarily focused on heart rate extraction, not the fidelity of the reconstructed PPG signal.

\textbf{Acknowledgments.} This material is in part based on work supported by the National Science Foundation CAREER Award under Grant No. 2338981.

\bibliographystyle{splncs04}
\bibliography{main}
\end{document}